\documentclass{article}
\usepackage{amsthm}

\usepackage{mathrsfs}
\usepackage{graphicx}
\usepackage{algorithm}
\usepackage{tabularx}

\newtheorem{lemma}{Lemma}

\usepackage{amsmath}
\usepackage{algorithmic}

\usepackage[preprint]{neurips_2025}
\usepackage{titletoc}

\usepackage[utf8]{inputenc} 
\usepackage[T1]{fontenc}    
\usepackage{hyperref}       
\usepackage{url}            
\usepackage{booktabs}       
\usepackage{amsfonts}       
\usepackage{nicefrac}       
\usepackage{microtype}      
\usepackage{xcolor}         
\usepackage{enumitem} 
\usepackage{amsmath} 
\usepackage{natbib}
\setcitestyle{numbers,square}
\setlist[itemize]{noitemsep, topsep=0pt}
\usepackage{diagbox} 
\usepackage{float}
\usepackage{tablefootnote}
\usepackage{multirow}
\usepackage{multicol}
\usepackage{adjustbox}
\usepackage{subfig}
\usepackage{graphicx}
\usepackage{caption}
\usepackage{longtable}

\usepackage{graphicx}
\usepackage{makecell}
\usepackage{wrapfig}

\title{One Step is Enough: Multi-Agent Reinforcement Learning Based on One-Step Policy Optimization for Order Dispatch on Ride-Sharing Platforms}

\author{%
  Zijian Zhao \\
  The Hong Kong University of Science and Technology \thanks{Corresponding Author: Sen Li}\\
  \And
  Sen Li \\
  The Hong Kong University of Science and Technology \\
  The Hong Kong University of Science and Technology (Guangzhou)
}

\begin{document}

\maketitle

\begin{abstract}

Order dispatch is a critical task in ride-sharing systems with Autonomous Vehicles (AVs), directly influencing operational efficiency and profitability. While Multi-Agent Reinforcement Learning (MARL) offers a scalable paradigm by decomposing the massive state-action space, existing methods are heavily reliant on accurate value function estimation. In large-scale, highly stochastic urban environments, such estimation is notoriously prone to bias and instability, severely limiting training efficiency and policy quality. 
To overcome this limitation, we propose two novel policy optimization methods that completely bypass the need for critic networks. First, we establish a formal connection between the homogeneity of AV fleets and the short-memory mixing property of the transportation network, proving that agent value functions converge to a common time-varying baseline up to a negligible residual. Leveraging this insight, we introduce Single-Trajectory Group Relative Policy Optimization (ST-GRPO), an adaptation of Large Language Models (LLMs) post-training techniques to multi-agent trajectories, which replaces the traditional value baseline with the fleet-wide average reward-to-go. Inspired by this reduction, we further derive One-Step Policy Optimization (OSPO), demonstrating that under the established structural priors, an optimal policy can be learned using only immediate, group-normalized rewards—rendering long-horizon bootstrapping unnecessary. 
Experiments on real-world ride-hailing datasets from Manhattan and Queens demonstrate that both ST-GRPO and OSPO achieve promising performance, particularly in reducing pickup times and increasing order service rates. Remarkably, both methods operate efficiently using simple Multilayer Perceptron (MLP) networks with minimal GPU utilization, underscoring the practical impact of exploiting domain structure for scalable MARL. 
Our code, trained models, and processed data are provided at the anonymous repository: \url{https://github.com/RS2002/OSPO}.

\end{abstract}

\section{Introduction}

The widespread adoption of on-demand ride-sharing platforms has fundamentally transformed urban transportation systems, offering scalable and sustainable mobility solutions for modern cities. By enabling multiple passengers with similar itineraries to share a single vehicle, ride-sharing platforms alleviate congestion, reduce vehicle kilometers traveled, and decrease urban fleet sizes, thereby addressing pressing challenges such as traffic, emissions, and inefficient resource utilization~\cite{wang2019ridesourcing,santi2014quantifying}. Recently, with the development of Autonomous Vehicle (AV) technology, systems like Robotaxi have been introduced by various companies, including Baidu, Didi, and Zoox. These systems not only offer lower operating costs through the AV fleet but also enable better management, as the fleet is fully controlled rather than relying on human drivers.

However, managing such platforms presents significant operational complexity. Passengers arrive randomly over time, each with unique origins and destinations, and the platform must dynamically determine not only how to bundle new orders together, but also how to assign each bundle to a suitable vehicle. These dispatching decisions must take into account the current routing of each vehicle, the destinations of onboard passengers, and the service quality requirements of all users. Critically, all decisions must be made in real time, under uncertainty regarding future demand and supply, making the order dispatch problem a central and challenging task for ride-sharing platforms with AVs.

Recently, Reinforcement Learning (RL) methods have shown great promise in addressing the order dispatch problem on ride-sharing platforms. Although order dispatch is fundamentally a centralized task, requiring the platform to aggregate information from all vehicles and orders to make globally optimal decisions, the massive state and action spaces encountered in realistic settings (which can reach millions or billions~\cite{sivagnanam2024multi}) pose a serious Curse of Dimensionality (CoD), rendering Single-Agent Reinforcement Learning (SARL) approaches intractable. As a result, most recent works adopt the Multi-Agent Reinforcement Learning (MARL) framework, where each vehicle is treated as an individual agent, and a global controller coordinates the agents to optimize the overall system objectives. MARL methods are typically categorized into three types: Decentralized Training with Decentralized Execution (DTDE), Centralized Training with Decentralized Execution (CTDE), and Centralized Training with Centralized Execution (CTCE)~\cite{jin2025comprehensive}. Most existing ride-sharing approaches follow the DTDE paradigm, since centralized methods still face scalability limitations similar to SARL, and are thus difficult to apply in large-scale scenarios with hundreds or thousands of agents. However, DTDE methods rely on the independent assumption that each agent acts based only on its local observation and treats other agents as part of the environment. This often leads to unstable training and poor cooperation among agents, as well as large estimation errors in Q-values and V-values~\cite{hu2025bmg}, ultimately resulting in suboptimal and inefficient learning. Due to page limitation, we left the detailed literature review in Appendix \ref{sec:related_work}.


To address this problem, we propose two novel multi-agent policy optimization methods that systematically exploit the homogeneous nature of AV fleets and the short-memory mixing properties of large-scale ride-sharing systems (formalized in Section \ref{sec:preliminary}). Crucially, these structural priors allow us to circumvent the notoriously difficult task of estimating Q-values or V-values in non-stationary multi-agent environments. 
First, we adapt Group Relative Policy Optimization (GRPO) \cite{GRPO} to the order dispatch task by replacing the V-value baseline in Proximal Policy Optimization (PPO) \cite{PPO} with the group-average reward-to-go. In this adaptation, we replace group rewards computed over multiple trajectories with multiple agent rewards derived from a single trajectory, leading to our method, Single-Trajectory GRPO (ST-GRPO). This modification significantly reduces computational resource requirements and minimizes bias by eliminating the estimation errors associated with neural network-based methods. Furthermore, inspired by the baseline replacement strategy, we introduce a more efficient method called One-Step Policy Optimization (OSPO), which can be trained using only one-step group rewards. By analyzing the one-step Bellman expansion under the established homogeneity and short-memory conditions, we prove that the advantage function reduces to the immediate, group-normalized reward (up to a negligible residual). Both methods are validated through case studies using real-world ride-hailing data and demonstrate promising performance with high efficiency. The main contributions of this paper are summarized as follows:

\begin{itemize}[left=0pt]
    \item We introduce Single-Trajectory GRPO, a novel order dispatch method for AV-based ride-sharing systems built upon GRPO. By replacing the PPO baseline with the group-average reward-to-go, our method removes the necessity of estimating V-values, thereby eliminating the estimation bias introduced by neural network critics. Moreover, by substituting group rewards computed over multiple trajectories with multi-agent rewards drawn from a single trajectory, ST-GRPO further enhances training efficiency.

    \vspace{0.1cm}
    \item We formally derive that, under the structural assumptions of fleet homogeneity and rapid mixing, the long-term advantage function collapses to the one-step group-relative reward. This finding establishes a new complexity lower bound for policy optimization within this class of large-scale cooperative MARL problems.

    \vspace{0.1cm}
    \item Extensive experiments on real-world datasets demonstrate that ST-GRPO and OSPO not only surpass existing DTDE, CTDE, and CTCE baselines in terms of service rate and pickup efficiency, but also achieve these gains using only simple Multilayer Perceptron (MLP) backbones with substantially lower GPU utilization. Notably, OSPO attains the best overall performance while maintaining the lowest computational footprint among all compared methods
\end{itemize}

\section{Preliminary} \label{sec:preliminary}
\subsection{Problem Setup}
In this paper, we study the optimal order assignment problem for a ride-sharing platform with $n$ homogeneous autonomous vehicles (AVs), which are treated as agents, each with capacity $c$. At each time step $t$, a set of $w_t$ new ride requests arrives from passengers at different locations. The platform must make real-time dispatch decisions by assigning each order to an appropriate vehicle. These decisions should account for the current route of each vehicle, the destinations of onboard passengers, and the service quality requirements of all users. We formulate this problem as a Multi-Agent Markov Decision Process (MAMDP) \cite{littman1994markov}, defined by $\langle n, S, U, \text{P}, R, \gamma \rangle$, where $n$ denotes the number of agents and $\gamma$ is the discount factor. The global state is defined as $S=(s^o, s_1, s_2, \ldots, s_n)$, where $s^o$ contains the information of unassigned orders and is observable to all agents, while $s_i$ denotes the state of agent $i$. For notational convenience, in the remainder of this paper we use $s_i$ to represent the concatenated state $[s_i, s^o]$. In our setting, $s^o$ includes the origins, destinations, and arrival times of all unconfirmed orders. The agent state $s_i$ includes the current location of vehicle $i$, its remaining capacity, its cumulative reward since the beginning of the episode, the average cumulative reward across all agents, and the information of all onboard orders, including their destinations, estimated total travel times, and estimated remaining travel times. The global action $U=(u_1,u_2,\ldots,u_n)$ is the joint action of all agents. At time step $t$, each agent action $u_i$ is represented as a $w_t$-dimensional binary vector, where $u_{i,j}=1$ indicates that order $j$ is assigned to agent $i$, and $w_t$ denotes the number of candidate orders at time $t$. The transition function $\text{P}(\cdot)$ is not modeled explicitly, as we consider a model-free reinforcement learning setting. Following \cite{hu2025bmg}, we define the reward function as:
\begin{equation}
\begin{aligned}
\text{r}(s_{i}, u_{i}) & =
\begin{cases}
\eta_1 + \eta_2 p^{in}_{i} - \eta_3 p^{out}_{i} - \eta_4 \chi_{i} - \eta_5 \rho_{i}, |u_{i}|=1 \\
0, |u_{i}|=0
\end{cases} 
\label{eq:reward_func}
\end{aligned}
\end{equation}
where $\eta_1$ to $\eta_5$ are non-negative weights representing the platform's valuation of each term, $p^{in}_{i}$ and $p^{out}_{i}$ represent the income from passengers and the vehicle operational costs, respectively. The variables $\chi_{i}$ and $\rho_{i}$ represent the number of en-route orders that will exceed their scheduled time and the additional travel time of all en-route orders when the assigned order is added to the scheduled route of agent $i$, respectively (\emph{Note that these variables depend on the state $s_i$ and the newly assigned order indicated by $u_i$}). This reward function is designed to comprehensively consider the interests of the platform and passengers, mimicking the operation of a real-world ride-sharing platform. It is important to emphasize that $p^{in}_{i}$ and $p^{out}_{i}$ are calculated based on the order distance and the additional travel distance for the agent, respectively. When calculating travel time, we will utilize the Traveling Salesman Problem (TSP) solution to optimize the agent's route. 
Since order dispatch is a fully cooperative task, the global reward can be expressed as the sum of rewards from each individual agent, written as $\text{R}(S,U) = \sum_{i=1}^{n} \text{r}(s_{i}, u_{i})$.

\subsection{Group Relative Policy Optimization (GRPO)}

With the success and popularity of Reinforcement Learning from Human Feedback (RLHF), RL-based LLM post-training has garnered increasing attention. This has also led to the emergence of many new RL methods, such as Direct Preference Optimization (DPO) \cite{rafailov2023direct}, Reinforce Leave-One-Out (RLOO) \cite{ahmadian2024back}, and GRPO \cite{GRPO}. Among these, GRPO is noted for its simple format, high efficiency, and promising performance, which are core factors contributing to the success of DeepSeekMath \cite{GRPO} and DeepSeek-V3 \cite{liu2024deepseek}. Serving as a PG \cite{sutton1999policy} based method, the optimization objective of GRPO has a similar format to PPO \cite{PPO}, expressed as:
\begin{equation}
\begin{aligned}
& \text{J}_{GRPO}(\theta) =  - \iota \text{KL}(\pi_\theta \| \pi_{ref}) + \\
& \mathbb{E}_{s \sim d^{\pi_{\theta^-}}, u \sim \pi_{\theta^-}(s)} \left[ \min \left\{ \frac{\pi_\theta(u|s)}{\pi_{\theta^-}(u|s)} \hat{\text{A}}_{\pi_\theta}(s,u),  \text{CLIP}\left(\frac{\pi_\theta(u|s)}{\pi_{\theta^-}(u|s)}, 1 - \epsilon, 1 + \epsilon' \right) \hat{\text{A}}_{\pi_\theta}(s,u) \right\} \right] \ ,
\end{aligned}
\label{eq:GRPO}
\end{equation}
where $\pi$ denotes the policy, $d^{\pi}$ denotes the state distribution under policy $\pi$, $\theta$ and $\theta^-$ represent the current network parameters and the parameters used for experience collection, respectively, $s$ and $u$ denote the state and action, $\pi_{ref}$ is the reference policy (i.e., the initial policy before post-training), serving as a normalization term to prevent the policy from diverging too much from the reference, $\hat{\text{A}}_{\pi_\theta}$ represents the advantage function of policy $\pi_\theta$, and $\iota,\epsilon,\epsilon'$ are hyper-parameters We follow the improvement in \cite{yu2025dapo}, which suggests setting a higher value for $\epsilon'$ than for $\epsilon$ to encourage exploration.

Aside from the KL divergence term, the core difference between GRPO and PPO lies in the definition of the advantage function. In PPO, the original advantage function is defined as:
\begin{equation}
\begin{aligned}
\text{A}_{\pi}(s_t, u_t) = \text{Q}_{\pi}(s_t, u_t) - \text{V}_{\pi}(s_t) \ ,
\end{aligned}
\label{eq:advantage}
\end{equation}
where the Q-value can be estimated using real trajectories, and the V-value can be estimated through a critic network. In contrast, GRPO proposes a simpler and more efficient advantage function given by:
\begin{equation}
\begin{aligned}
\hat{\text{A}}_{\pi}(s_t, u_t) = \sum_{\tau \in K_t} \gamma^\tau \frac{\text{r}(s_{t+\tau}, u_{t+\tau}) - \mu}{\sigma} \ ,
\end{aligned}
\label{eq:grpo_advantage}
\end{equation}
where $\gamma$ is discount factor\footnote{$\gamma$ is set as $1$ in GRPO paper \cite{GRPO}.}, $K_t = \{0, 1, 2, \ldots, T-t\}$, $T$ represents the time horizon, $\text{r}(\cdot,\cdot)$ denotes the reward function, and $\mu$ and $\sigma$ represent the mean and standard deviation of the rewards across multiple trajectories with the same initial state and policy.

\subsection{Homogeneous and Short-Memory Properties in AV Ride-Sharing Systems} 

We apply GRPO to ride-sharing platforms with autonomous vehicles, where the method can be further tailored by exploiting two key properties of the problem setting: (i) homogeneity and (ii) short memory.

\textbf{(i) Homogeneity:} All AVs in the fleet are homogeneous, sharing the same vehicle model and characteristics, such as capacity and speed. Moreover, because all agents have identical state and action spaces, the optimal policy is expected to be the same across agents under the total cooperative game \cite{terry2020revisiting, sanjari2023optimality, sanjari2024optimality}. In particular, \cite{terry2020revisiting} showed that, for homogeneous agents, policy sharing during training can still converge to an optimal solution. Accordingly, in this paper, all agents share a common policy for order dispatch.

\textbf{(ii) Short-Memory:} We consider a short-memory ride-sharing system where the current state and action of each agent have limited influence on its decisions and rewards over long time horizons. Because travel demand exhibits strong temporal variation (e.g., rush hours vs.\ off-peak), the environment is inherently non-stationary. We capture the short-memory property in this non-stationary regime by requiring that the system mixes much faster than the timescale at which demand characteristics change appreciably. Formally, let $\mathcal{M}_t$ denote the exogenous demand parameters at time $t$ (origin-destination flows, etc.). We assume there exists a time window length $\tau_{\mathrm{mix}}$ such that over any interval of duration $\tau_{\mathrm{mix}}$, the variation of $\mathcal{M}_t$ is negligible, and within such an interval the state process of each agent satisfies a uniform $\beta$-mixing condition. To formalize this, let $\{s_{i,t}\}_{t\ge 0}$ be the state process of agent $i$ under policy $\pi$, and let $\mathcal{F}_{-\infty}^{t}$ and $\mathcal{F}_{t+k}^{\infty}$ be the $\sigma$-algebras generated by the process up to time $t$ and from time $t+k$ onward, respectively. The $\beta$-mixing coefficient at lag $k$ is defined as:
\begin{equation}
\beta_{\pi}(k) = \sup_{t} \; \frac{1}{2} \sup \sum_{i=1}^{I}\sum_{j=1}^{J} \bigl| \text{P}(A_i \cap B_j) - \text{P}(A_i)\text{P}(B_j) \bigr|,
\label{eq:beta_mixing_def}
\end{equation}
where the inner supremum is taken over all finite partitions $(A_i)$ of $\mathcal{F}_{-\infty}^{t}$ and $(B_j)$ of $\mathcal{F}_{t+k}^{\infty}$. The process is said to be uniformly $\beta$-mixing if $\beta_{\pi}(k) \to 0$ as $k \to \infty$. This property aligns with real ride-sharing systems: regardless of an AV's current location or en-route orders, after a period ranging from minutes to hours, the system's stochastic order dynamics will drive the AV into an arbitrary random status, independent of its initial condition.

In our non-stationary setting, the supremum over $t$ guarantees that the rate of memory decay is bounded independently of the specific time of day. Consequently, for any $\epsilon>0$, there exists a finite lag $\kappa(\epsilon)$ such that $\beta_{\pi}(\kappa) \le \epsilon$, uniformly in $t$.

For Markovian state processes (which holds under policy $\pi$), a standard consequence of uniform $\beta$-mixing is that the total variation distance between the $\kappa$-step state distributions starting from any two initial states $s, s'$ is bounded by $\epsilon$, irrespective of the starting time:
\begin{equation}
\forall \epsilon>0,\; \exists \kappa>0,\; \sup_{t} \sup_{s,s'} \bigl\| \text{P}_{\pi}(s_{t+\kappa}\mid s_{t}=s) - \text{P}_{\pi}(s_{t+\kappa}\mid s_{t}=s') \bigr\|_{\mathrm{TV}} \le \epsilon.
\label{eq:short_memory_tv}
\end{equation}
This property relies on two operational conditions: (a) The operational region is sufficiently open, allowing orders to originate from and be destined to arbitrary locations without being constrained by region-specific isolation. (b) The policy ensures that no vehicle is restricted to a particular operational region. Prior work \cite{sun2022optimizing,sun2024optimizing} demonstrates that a well-designed policy naturally satisfies this condition, with typical mixing times on the order of minutes, which is much shorter than the timescale over which demand profiles shift (tens of minutes to hours). Hence the uniform $\beta$-mixing assumption is empirically justified.

\subsection{Value Homogeneity Property in AV Ride-Sharing Systems}
Leveraging the homogeneity and short-memory properties, we identify a value homogeneity property among agents, formally defined as follows: there exists a time-dependent fleet-wide mean V-value $\bar{V}_t$ and small residual terms $\omega(\cdot)$ such that for every agent $i \in \mathcal{I} = \{1,\dots,n\}$,
\begin{equation}
\text{V}_{\pi}(s_{i,t}) = \bar{V}_t - \omega(s_{i,t}), \qquad \text{where } \omega(s_{i,t}) \ll \bar{V}_t.
\label{eq:assumption}
\end{equation}
The residual is defined as the deviation of the agent's value from the baseline $\bar{V}_t$:
\begin{equation}
\omega(s_{i,t}) := \bar{V}_t - \text{V}_{\pi}(s_{i,t}), \quad \text{with} \quad \bar{V}_t := \mathbb{E}_{i\in \mathcal{I},\, s_{i,t} \sim d^\pi_t(i)}\bigl[\text{V}_{\pi}(s_{i,t})\bigr],
\end{equation}
where $d^\pi_t(i)$ denotes the state distribution of agent $i$ under policy $\pi$ at time step $t$. Note that $\omega$ and $\bar{V}$ implicitly depend on $\pi$, but we omit this dependence for notational simplicity. A detailed proof is provided in Appendix \ref{sec:assumption}, and the conditions under which our methods satisfy these properties are discussed in Appendix \ref{sec:condition}.

\section{Methodology}

\begin{figure*}[htbp]
\centering 
\includegraphics[width=0.9\textwidth]{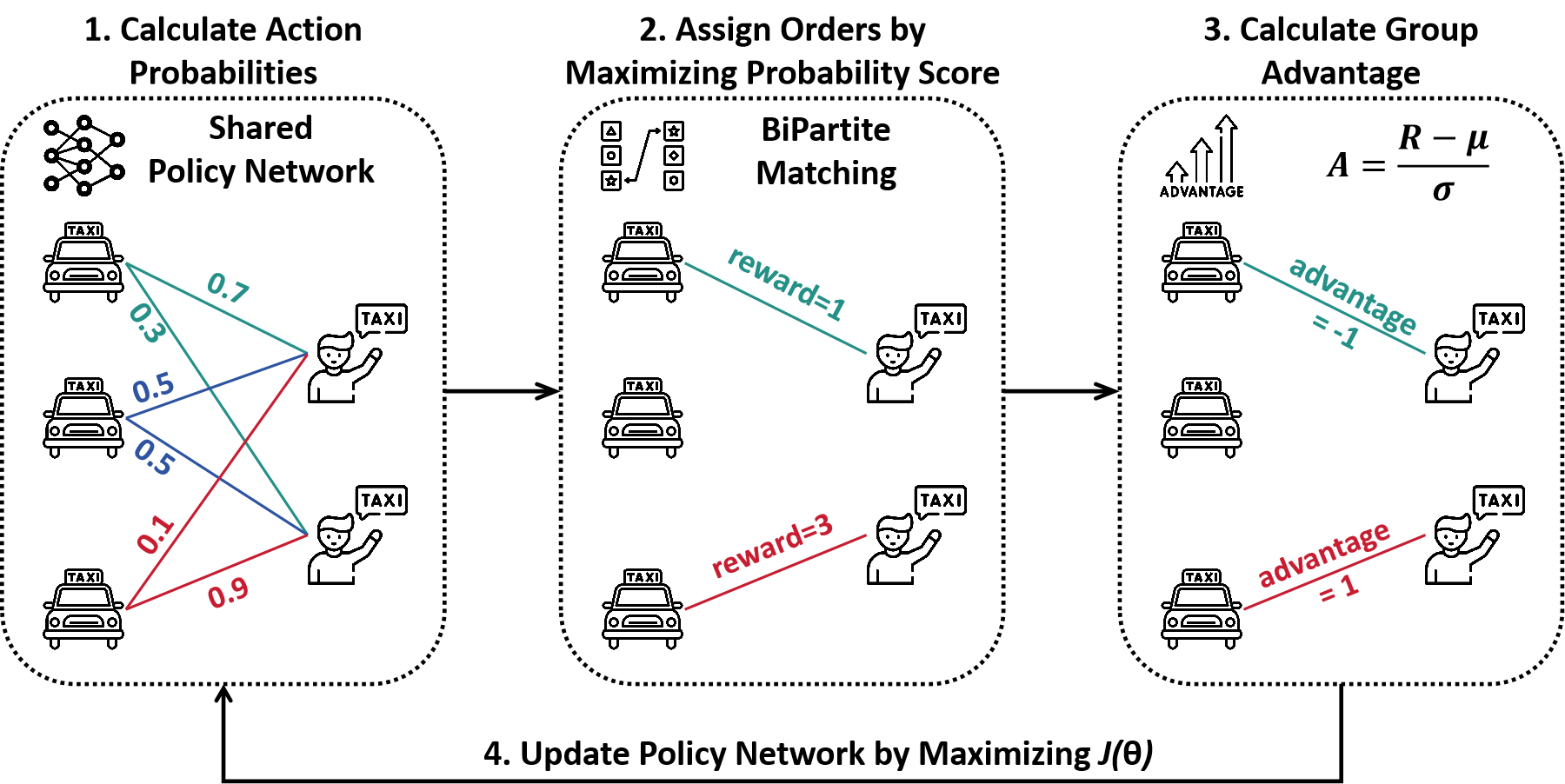}
\caption{The training process consists of four steps: (1) Calculate the matching probability for each vehicle-order pair; (2) Assign orders by maximizing the probability score; (3) Calculate the group advantage; (4) Train the network using policy gradients.}
\label{fig:main}
\end{figure*}


\subsection{Policy-Based Method for Order Dispatch in Ride-Sharing System}

In policy-based deep RL, at each time step $t$, the agent directly generates the probabilities of each action based on the observation:
\begin{equation}
\begin{aligned}
p_{i,t} = \pi_{\theta}(s_{i,t}) \ ,
\label{eq:action}
\end{aligned}
\end{equation}
where $s_{i,t}$ represent the state of agent $i$ at time $t$, and $p_{i,t} \in [0,1]^{w_t}$ represents the action probability (i.e., the probability of matching each order to agent $i$). Due to the homogeneity of the agents, we utilize a shared network among all agents, similar to approaches taken in previous works.

However, in the order dispatch task, we cannot assign the same order to multiple agents, making it infeasible to directly sample actions as $u_{i,t} \sim p_{i,t}$. To address this problem, we follow previous policy-based ride-sharing methods \cite{enders2023hybrid, hoppe2024global} by viewing the probabilities as scores and using bipartite matching to maximize the assignment score:
\begin{subequations} \label{eq:ILP} 
\begin{align} 
& \max_{U_t} \sum_{i \in \mathcal{I}} u_{i,j,t} \cdot y_{i,j,t}, \label{match_obj} \\ 
\text{s.t.} \quad 
& \sum_{i \in \mathcal{I}} u_{i,j,t} \leq 1, \quad \forall j \in \mathcal{J}_t, \label{match_order} \\ 
& \sum_{j \in \mathcal{J}_t} u_{i,j,t} \leq 1, \quad \forall i \in \mathcal{I}, \label{match_driver} \\ 
& u_{i,j,t} \in \{0,1\}, \quad \forall i \in \mathcal{I}, j \in \mathcal{J}_t, \label{constraint}
\end{align} 
\end{subequations}
where $u_{i,j,t}$ indicates whether agent $i$ is assigned to order $j$ at time $t$ (with 1 indicating assignment and 0 indicating no assignment), and $y_{i,j,t}$ denotes the score $p_{i,j,t}$ of agent $i$ choosing order $j$ at time $t$ (with $y_{i,j,t} = -\infty$ for all unavailable workers at time $t$). The set $\mathcal{I} = \{1,2,\ldots,n\}$ represents the agents, and the set $\mathcal{J}_t = \{1,2,\ldots,w_t\}$ represents the orders to be assigned. Constraint (\ref{match_order}) ensures that an order can be assigned to at most one agent, while constraint (\ref{match_driver}) guarantees that each agent is assigned at most one order. This process is illustrated in the first two steps of Fig. \ref{fig:main}.

\subsection{Single-Trajectory GRPO (ST-GRPO) Solution}

In the original GRPO \cite{GRPO}, the advantage function is defined as in Eq. \eqref{eq:grpo_advantage}, which eliminates the need for a critic network and improves training efficiency. Although GRPO has proven effective in LLM fine-tuning across numerous studies, adapting it directly to large-scale ride-sharing systems presents two notable limitations. (i) First, the advantage formulation in Eq. \eqref{eq:grpo_advantage} lacks a clear theoretical connection to the standard advantage definition in Eq. \eqref{eq:advantage}, making its interpretation less transparent. (ii) Second, GRPO relies on repeatedly sampling from the same initial state (i.e., the same prompt in NLP). In a ride-sharing simulator with hundreds of thousands of AV agents, such repeated sampling is computationally prohibitive. Moreover, in practical deployments, resetting the entire system to an identical initial state of all agents and orders is infeasible. To overcome these limitations, we propose an adapted GRPO method that operates on a single trajectory collected from all agents. Our derivation starts from the standard advantage definition (Eq. \eqref{eq:advantage}) and naturally leads to a form reminiscent of the original GRPO, as detailed below.

Under policy $\pi$, we collect a single system trajectory $\mathcal{T}^\pi = \{S_0,U_0,R_0,S_1,U_1,R_1,\dots\}$ involving all agents. For agent $i$, let $\mathcal{T}^\pi_{i,t}$ denote the sub-trajectory from time step $t$ onward, and define its reward-to-go as
\begin{equation}
G(\mathcal{T}^\pi_{i,t}) = \sum_{\tau=t}^{T} \gamma^{\tau-t} r_{i,\tau},
\label{eq:reward_to_go}
\end{equation}
where $G(\mathcal{T}^\pi_{i,t})$ can serve as an unbiased estimator of the state-action value $Q_{\pi}(s_{i,t}, u_{i,t})$ (i.e. $\hat{Q}_{i,t}=G(\mathcal{T}^\pi_{i,t})$).

The V-value homogeneity property (Eq. \eqref{eq:assumption}) states that all agents share a common time-dependent baseline $\bar{V}_t$ up to a small residual $\omega_{i,t}$. Expanding $V_{\pi}(s_{i,t})$ as an expectation over trajectories and then taking the population mean yields
\begin{equation}
\begin{aligned}
\text{V}_{\pi}(s_{i,t}) & = \mathbb{E}_{i \in \mathcal{I}}\bigl[ \text{V}_{\pi}(s_{i,t}) \bigr] - \omega(s_{i,t}) \nonumber = \mathbb{E}_{i \in \mathcal{I}}\Bigl[ \mathbb{E}_{\mathcal{T}^\pi_{i,t}\sim \rho_t^{\pi}(i)}\bigl[ G_{i,t} \bigr] \Bigr] - \omega(s_{i,t}) \nonumber \\
& = \mathbb{E}_{i \in \mathcal{I},\, \mathcal{T}^\pi_{i,t}\sim \rho_t^{\pi}(i)}\bigl[ G_{i,t} \bigr] - \omega(s_{i,t}) \approx \mathbb{E}_{i \in \mathcal{I},\, \mathcal{T}^\pi_{i,t}\sim \rho_t^{\pi}(i)}\bigl[ G_{i,t} \bigr] ,
\label{eq:V_estimation}
\end{aligned}
\end{equation}
where $\rho_t^{\pi}(i)$ represents the sub-trajectory distribution of agent $i$ from step $t$ under policy $\pi$.

Eq. \eqref{eq:V_estimation} reveals that the desired value $\text{V}_{\pi}(s_{i,t})$ equals the expectation of the reward-to-go over both the fleet distribution and the trajectory randomness, minus a small bias. Since we only have access to a single system trajectory, we approximate this double expectation by its sample mean across the fleet:
\begin{equation}
\hat{V}_t = \frac{1}{|\mathcal{I}|} \sum_{i\in\mathcal{I}} G_{i,t}.
\label{eq:fleet_baseline}
\end{equation}
This substitution is justified by viewing $\hat{V}_t$ as a Monte Carlo estimator of $\mathbb{E}_{i \in \mathcal{I},\, \mathcal{T}^\pi_{i,t}\sim \rho_t^{\pi}(i)}[G_{i,t}]$ with $|\mathcal{I}|$ samples, where each agent contributes one draw from its own trajectory distribution. In a large fleet with fast mixing dynamics, the returns $G_{i,t}$ exhibit limited mutual dependence, and their sample mean $\hat{V}_t$ concentrates sharply around its expectation $\bar{V}_t$. Under such a circumstance, we can also get the estimation of V-value of each agent $\hat{V}_{i,t}\approx \hat{V}_t$.

Consequently we can write an unbiased estimation of advantage function as:
\begin{equation}
\begin{aligned}
 &  \hat{\text{A}}^{GRPO}_{\pi}(s_{i,t}, u_{i,t}) = \frac{1}{\sigma_{t:}} \left( \hat{Q}_{i,t} - \hat{V}_{i,t} \right) \approx \frac{1}{\sigma_{t:}} \left( G_{i,t} - \mu_{t:} \right) \ , \\
& \mu_{t:}    = \hat{V}_t , \quad \sigma_{t:}  = \sqrt{\mathbb{E}_{i \in \mathcal{I}} \left[ (G_{i,t} - \mu_{t:})^2\right]} \ .
\end{aligned}
\label{eq:grpo_advantage3}
\end{equation}
where $\sigma_{t:}$ is utilized for normalization to improve training efficiency, utilized by many works \cite{GRPO, kurin2022defense}. 

Additionally, when using the target function in Eq. \eqref{eq:GRPO} within the MARL scenario, we note that there is no reference policy $\pi_{ref}$ as found in LLMs. We innovatively propose regularly tracking policy performance and retaining the best checkpoint. This best policy will serve as the reference policy, helping to ensure that the current policy does not deviate significantly from the optimal one, aided by the KL normalization term.

\subsection{One Step Policy Optimization (OSPO) Solution}

Based on the homogeneous property in Eq. \eqref{eq:assumption}, we can recalculate the advantage function as follows:
\begin{equation}
\begin{aligned}
& \text{A}^{OSPO}_{\pi}(s_{i,t},u_{i,t})  =  \text{Q}_{\pi}(s_{i,t}, u_{i,t}) - \text{V}_{\pi}(s_{i,t}) \\
= & \underbrace{\text{r}(s_{i,t}, u_{i,t}) - \mathbb{E}_{i \in \mathcal{I}, s_{i,t} \sim d^\pi_t(i), u_{i,t} \sim \pi(s_{i,t})} [\text{r}(s_{i,t}, u_{i,t}))]}_{\text{Reward Term}} \underbrace{- \gamma \mathbb{E}_{s_{i,t+1} \sim \text{P}(\cdot|s_{i,t}, u_{i,t})} [\omega(s_{i,t+1})] + \omega(s_{i,t})}_{\text{Residual Term}}, \\
\end{aligned}
\label{eq:marl_advantage}
\end{equation}
In Appendix \ref{sec:ospo}, we prove that when the $\overline{V}_t$ has an upper bound, which holds when the horizon is limited by $T$ and the mixing time is far less than $T$, one can select an appropriate discount factor $\gamma$ and mixing lag $\kappa$ to make the residual term arbitrarily small and thus negligible.

Then, if we perform a similar operation to GRPO, using the average reward to replace the expectation, applying reward normalization \cite{kurin2022defense}, and ignoring the residual term, we arrive at the advantage function:
\begin{equation}
\begin{aligned}
& \hat{\text{A}}^{OSPO}_{\pi}(s_{i,t},u_{i,t})  = \frac{\text{r}(s_{i,t}, u_{i,t}) - \mu_t}{\sigma_t} , \\
& \mu_{t} = \frac{1}{|\mathcal{I}|}\sum_{i \in \mathcal{I}}\text{r}(s_{i,t}, u_{i,t}) , \quad \sigma_{t}  = \sqrt{\mathbb{E}_{i \in \mathcal{I}} \left[ (\text{r}(s_{i,t}, u_{i,t}) - \mu_{t})^2\right]} \ .
\end{aligned}
\label{eq:new_advantage2}
\end{equation}


\subsection{A Heuristic Regulation Term}

In this paper, our proposed ST-GRPO and OSPO are both based on the approximate V-value homogeneity property in Eq. \eqref{eq:assumption}. However, this property may not always hold during the training exploration phase, which could lead to incorrect update directions. To address this issue, we propose adding a heuristic regulation term to better align with this property. Intuitively, the most straightforward approach is to collect the full trajectory, compute the deviation of reward-to-go among agents, and use it as a penalty term for the reward or advantage function. However, collecting the full trajectory would break the efficiency of OSPO, which relies on only one-step samples. Therefore, we propose to replace it with the change in historical cumulative reward deviation. (Note that if agents have similar historical cumulative rewards at any time, they also have similar V-values at any time.) Consequently, the final advantage function is expressed as:
\begin{equation}
\begin{aligned}
\hat{\text{A}}_{\pi}'(s{i,t},u_{i,t}) & = \hat{\text{A}}_{\pi}(s_{i,t},u_{i,t}) - \alpha (\zeta_{t} - \zeta_{t-1}) , \\
\zeta_{t} & = \operatorname{Std}\left( \{ \sum_{\tau=0}^t \text{r}(s_{i,\tau},u_{i,\tau}) \mid i \in \mathcal{I} \} \right),
\end{aligned}
\label{eq:new_advantage3}
\end{equation}
where $\alpha$ is a hyper-parameter and $\hat{\text{A}}$ is the advantage function of OSPO and ST-GRPO. A detailed algorithm description is provided in Appendix \ref{sec:alg}. Note that when Eq. \eqref{eq:assumption} holds, the regulation term is close to zero, thus having limited impact on the optimal policy.


\section{Experiment}
\subsection{Experiment Setup} \label{sec:setup}

To validate our method, we conduct a series of experiments using a real-world ride-hailing dataset from Manhattan, New York City \cite{TLCData}. The data collected from 19:00 to 19:30 on July 17, 2024, is used as the training set, which includes 3,726 valid orders. We then evaluate the trained model using data from the entire day of July 18, 2024, selecting the first 30 minutes of data from each hour, resulting in a total of 24 independent tests. An expanded generalization experiment is provided in Appendix \ref{sec:expand}, using the data from Queens, New York.

For the simulation, we set the total number of vehicles to 1,000, with each car having a capacity of 3 and a default speed of 60 km/h. The optimal routing is achieved using the OSRM simulator \cite{luxen-vetter-2011}. The experiments are conducted using the PyTorch framework \cite{paszke2019pytorch} on a workstation running Windows 11, equipped with an Intel(R) Core(TM) i7-14700KF processor and an NVIDIA RTX 4080 graphics card. Following the setup of pervious works, we set each episode length to 30 minutes, with each step representing 1 minute, resulting in a running time of approximately 40 to 120 seconds for different methods per episode. We maximize the training episodes at 1,000 and evaluate model performance every 10 episodes. 

To further illustrate the efficiency of our method, we choose a simple four-layer MLP as the policy network, with 128 units in each hidden layer and LeakyReLU \cite{maas2013rectifier} as the activation function. Considering the variable action space (since the number of orders changes each time), we set the network input as a single vehicle-order pair, ignoring the relationships between orders. Finally, a softmax layer is used to normalize the output probabilities. For exploration, we add gradually decreasing random noise, employing the same Binary Symmetric Channel (BSC) noise as in \cite{hu2025bmg,hoppe2024global,enders2023hybrid,al2019deeppool}.

In the following, we first compare our ST-GRPO and OSPO ride-sharing methods with previously popular methods of different types, including DTDE, CTDE, and CTCE. In Appendix \ref{sec:ablation}, we conduct a series of ablation studies to detail the effects of different modules in ST-GRPO and OSPO, comparing them with similar policy gradient-based methods, including Independent PPO (IPPO) \cite{de2020independent}, Multi-Agent PPO (MAPPO) \cite{yu2022surprising}, and Independent PG (IPG) \cite{sutton1999policy}. All comparisons are conducted under identical conditions to ensure fairness. For optimization, we use the Adam optimizer \cite{kingma2014adam} with an initial learning rate of $10^{-4}$, a decay rate of $0.99$, and a batch size of $256$. For MDP, we set the discount factor $\gamma$ as $0.99$.

Finally, we report the hyper-parameters used in our method. The weight for the KL term, denoted $\iota$, is set to $0.5$, and the weight for the heuristic term, denoted $\alpha$, is set to $1$. Following \cite{yu2025dapo}, the CLIP parameters $\epsilon$ and $\epsilon'$ are set to $0.2$ and $0.28$, respectively. 

\begin{table*}[htbp]
    \caption{Comparison of Different Ride Sharing Methods: \textbf{Bold} entries represent the best results.}
    \centering
    \begin{adjustbox}{width=1.00\textwidth}
    \begin{tabular}{l@{\quad}|cc|cccccc}
        \toprule
        \textbf{Method} & \textbf{OSPO} & \textbf{ST-GRPO} & \textbf{DeepPool} & \textbf{BMG-Q}  & \textbf{HIVES}  & \textbf{Enders et al.}  & \textbf{CEVD} & \textbf{Hoppe et al.} \\
        \midrule
        \textbf{Reward Type} & Local & Local & Local & Local & Global & Local & Global & Mixed \\
        \textbf{RL Algorithm} & OSPO & ST-GRPO & IDDQN  & IDDQN & QMIX  & MASAC & VD & COMA  \\
        \textbf{Network Backbone} & MLP & MLP &  MLP & GAT & GRU & MLP & MLP & MLP \\ 
        \midrule
        \textbf{Model Size} & \textbf{20K} & \textbf{20K} & \textbf{20K} & 117K & 16M & 118K & 23K & 167K \\
        \textbf{GPU Occupation (GB)} &  \textbf{3.82} & 5.17 & 3.97 & 4.28 & 6.01 & 8.19 & 21.45 & 9.78 \\
        \textbf{Time per Episode (s)} & 30.22  & 33.75 & 35.36 & 62.76 & 69.17 & 91.82 & 80.44 & 192.70 \\
        \midrule
        \textbf{Reward ($\times10^3$)} & \textbf{11.00 ± 2.63} &10.90 ± 2.61 & 10.41 ± 2.34 & 10.88 ± 2.35 & 10.70 ± 2.32 & 9.77 ± 2.41& 10.71 ± 2.36 &9.93 ± 2.43 \\ 
        \textbf{Service Rate} &\textbf{1.00 ± 0.00} &\textbf{1.00 ± 0.00} & 0.99 ± 0.02& 1.00 ± 0.01&0.99 ± 0.02 &0.99 ± 0.02 &1.00 ± 0.01 & 0.99 ± 0.02\\ 
        \textbf{Delivery Time} &12.32 ± 0.71 & 12.40 ± 0.73& 12.20 ± 1.43& \textbf{10.61 ± 0.70}& 10.81 ± 0.51& 12.65 ± 0.52&11.95 ± 0.79 &12.54 ± 0.61 \\ 
        \textbf{Detour Time} &1.26 ± 0.41 & 1.42 ± 0.52& 1.24 ± 0.90& \textbf{0.59 ± 0.59}& 0.84 ± 0.79& 2.72 ± 0.33&  1.12 ± 0.85&2.54 ± 0.43 \\ 
        \textbf{Pickup Time} & \textbf{6.71 ± 0.64} & 7.33 ± 0.60& 8.67 ± 1.23& 7.62 ± 0.23& 8.13 ± 0.31& 8.05 ± 0.18&8.02 ± 0.88 & 8.10 ± 0.58\\ 
        \textbf{Confirmation Time} & \textbf{0.00 ± 0.00}& \textbf{0.00 ± 0.00}& 0.01 ± 0.03& 0.00 ± 0.02& 0.00 ± 0.01& 0.06 ± 0.20& 0.00 ± 0.01& 0.02 ± 0.08\\ 
        \bottomrule
    \end{tabular}
    \end{adjustbox}
    \label{tab:result}
\end{table*}

\subsection{Experiment Results}

\begin{figure*}[t!]
\centering 
\includegraphics[width=0.95\textwidth]{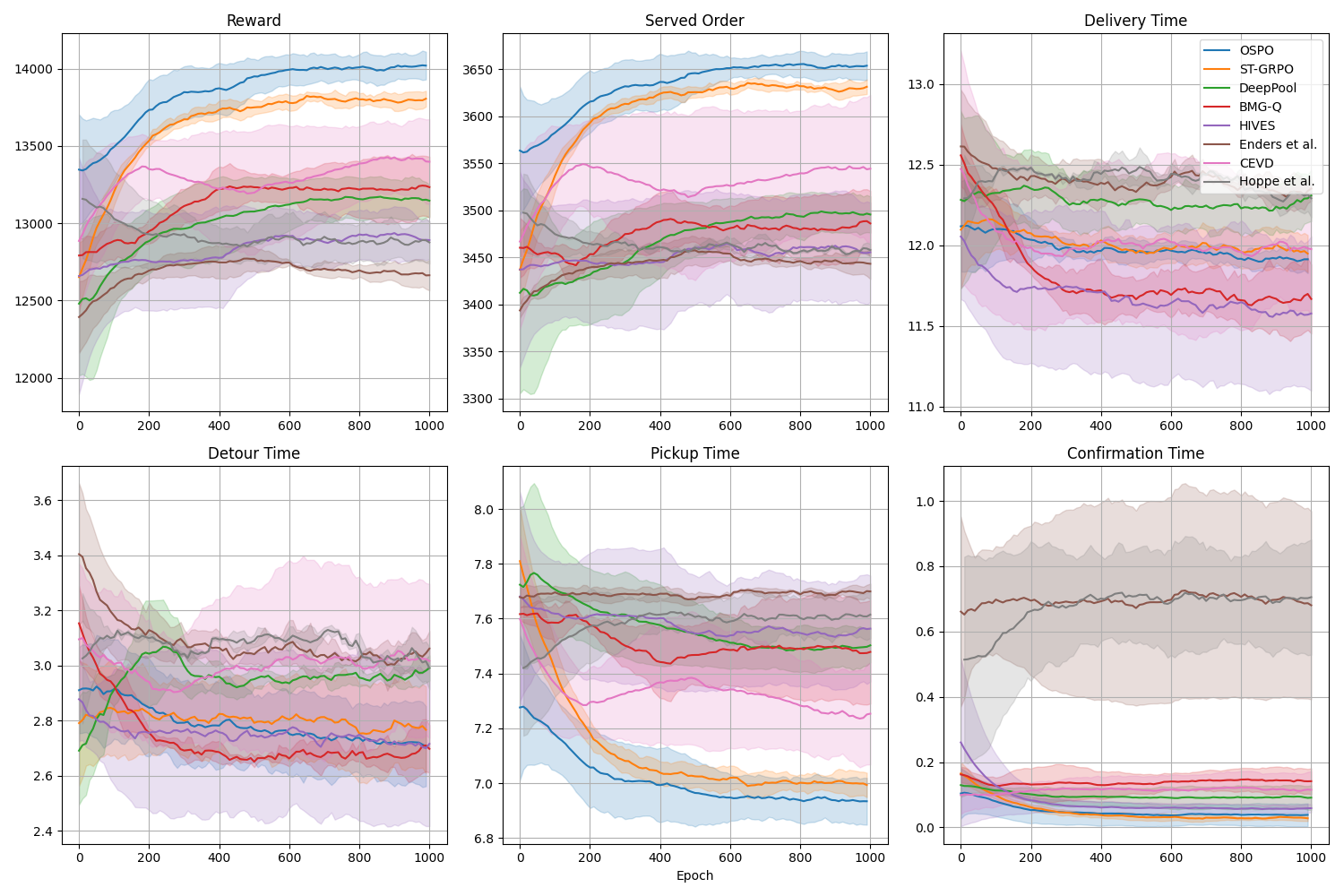}
\caption{Training Process: Each method is trained five times. The shaded area represents the standard deviation of each method.}
\label{fig:training}
\end{figure*}

The training process and testing results are presented in Fig. \ref{fig:training} and Table \ref{tab:result}, respectively. We illustrate the cumulative reward alongside the number of orders served, as well as the average delivery time, detour time, pickup time, and confirmation time for each order. Detailed explanations of these metrics can be found in Appendix \ref{sec:metric}. According to the results, we observe that ST-GRPO and OSPO outperform the others primarily by achieving lower pickup and confirmation times, which allows them to serve more orders. Notably, OSPO, with the smallest network size and lowest GPU utilization, achieves the best performance across most testing scenarios. Please note that the results in Table \ref{tab:result} represent the weighted average values from 24 different episodes, calculated according to order volume, along with the standard deviation. More details, including a comprehensive ablation study and additional experiments in another scenario, can be found in Appendix~\ref{sec:detail}.

\section{Conclusion}


In this paper, we introduced ST-GRPO and OSPO, two novel MARL algorithms tailored for the large-scale order dispatch problem in AV ride-sharing platforms. By exploiting the inherent homogeneity and short-memory mixing properties of AV fleets, we formally demonstrated that the value functions across agents converge to a common time-varying baseline, enabling a radical simplification of the advantage estimation process. ST-GRPO successfully adapts the group-relative paradigm from language model fine-tuning to the multi-agent trajectory setting, while OSPO pushes this reduction to its logical extreme, demonstrating that one step is indeed enough: competitive policies can be learned using only immediate, group-normalized rewards without any critic network or long-horizon backpropagation. Our extensive experiments on real-world Manhattan and Queens datasets confirm that both methods not only match but often surpass the performance of competitive MARL baselines, achieving lower pickup times and higher service rates, all while utilizing simple MLP backbones with low GPU memory. More discussion could be found at Appendix \ref{sec:discussion}.



\bibliographystyle{ieeetr}
\bibliography{ref.bib}


\appendix


\newpage
\section*{Appendix Contents}  
\startcontents  
\printcontents{}{1}{\setcounter{tocdepth}{2}}  
\newpage

\section{Related Work} \label{sec:related_work}
In this section, we provide a brief overview of current ride-sharing methods. Starting from the Independent Double Deep Q-Learning (IDDQN)-based DeepPool \cite{al2019deeppool}, DTDE methods have become a mainstream paradigm, as the large number of agents, along with the extensive state and action spaces, severely hinders the development of centralized methods. Subsequent works have primarily aimed to enhance algorithms by fostering better cooperation among agents. For example, Graph Neural Networks (GNNs) have been widely explored to effectively utilize neighboring information \cite{hu2025bmg}. \cite{li2019efficient} also propose using mean field approaches to capture more useful global information.

Additionally, some works attempt to transfer the successes of CTDE and CTCE to ride-sharing, where centralized critics can facilitate better cooperation among agents. However, these approaches face various shortcomings and challenges. \cite{de2020efficient} first introduce QMIX to ride-sharing, but only consider tasks involving a limited number of vehicles in simulated grid scenarios, which presents a significant gap to real-world applications. \cite{hao2022hierarchical} explore hierarchical structures to address the curse of dimensionality in QMIX \cite{rashid2020monotonic}, but the input dimensions of the mixture network still increase with the number of agents. \cite{bose2023sustainable} propose a Value Decomposition (VD) \cite{sunehag2018value} based method; however, the credit assignment challenge becomes more complex in ride-sharing scenarios with hundreds of agents. 

Lastly, there exists a series of works aimed at improving system efficiency by jointly considering relocation \cite{zhang2024joint}, order bundling \cite{jiang2025optimizing}, and multi-modal transportation \cite{hu2025coordinating}. Since these topics extend beyond the scope of this paper, we do not discuss them further here.

Nevertheless, most methods rely on precise value-function estimation (e.g., Q/V-networks), which introduces bias and instability due to approximation errors \cite{hu2025bmg}. Our work addresses this gap by proposing ST-GRPO/OSPO—eliminating value estimation entirely through group rewards—for stable and scalable coordination in homogeneous MARL systems.

\section{Proof Derivation} \label{sec:proof}
\subsection{Approximate Value Homogeneity} \label{sec:assumption}
Under the shared policy $\pi$ and the uniform short-memory property, we now show that the value functions of all agents can be made arbitrarily close in a relative sense, provided the mixing is sufficiently fast and the discount factor is chosen appropriately. This result justifies the use of a common time-varying baseline in our subsequent decentralized coordination.

Because the environment is non-stationary, the value function of an agent starting from state $s$ at time $t$ is defined as
\begin{equation}
\text{V}_{\pi}(s; t) = \mathbb{E}_{\pi}\Bigl[ \sum_{\tau=0}^{\infty} \gamma^{\tau} r(s_{t+\tau}, u_{t+\tau}) \mid s_{t}=s \Bigr],
\end{equation}
where $\text{r}(s,u) \in [0, R_{\max}]$ is the bounded one-step reward (the agent can always choose to take no order and receive zero reward, so non-negativity is natural). The discount factor $\gamma \in (0,1)$ is fixed. Although $\text{V}_{\pi}(\cdot; t)$ depends on time, the uniform mixing condition \eqref{eq:short_memory_tv} ensures that the difference between values of any two states can be uniformly controlled.

Fix an arbitrary $\epsilon>0$. By the uniform short-memory property, choose $\kappa$ such that the total variation bound \eqref{eq:short_memory_tv} holds for all $t$. For any two states $s, s'$ and any fixed time $t$, we decompose the value functions using the $\kappa$-step Bellman expansion:
\begin{align}
\text{V}_{\pi}(s; t) &= \mathbb{E}_{\pi}\Bigl[ \sum_{\tau=0}^{\kappa-1} \gamma^{\tau} r_{t+\tau} \mid s_t=s \Bigr] + \gamma^{\kappa} \mathbb{E}_{s_{t+\kappa}\sim \text{P}_{\pi}^{\kappa}(\cdot\mid s, t)}\bigl[ V_{\pi}(s_{t+\kappa}; t+\kappa) \bigr],
\end{align}
where $r_{t+\tau} = \text{r}(s_{t+\tau}, u_{t+\tau})$ and $\text{P}_{\pi}^{\kappa}(\cdot\mid s, t)$ denotes the $\kappa$-step state distribution starting from $s$ at time $t$. Their difference satisfies:
\begin{align}
\bigl| \text{V}_{\pi}(s; t) - \text{V}_{\pi}(s'; t) \bigr| &\le \underbrace{\Bigl| \mathbb{E}_{\pi}\bigl[ \sum_{\tau=0}^{\kappa-1} \gamma^{\tau} r_{t+\tau} \mid s_t=s \bigr] - \mathbb{E}_{\pi}\bigl[ \sum_{\tau=0}^{\kappa-1} \gamma^{\tau} r_{t+\tau} \mid s_t=s' \bigr] \Bigr|}_{\text{Term I}} \nonumber \\
&\quad + \gamma^{\kappa} \underbrace{\Bigl| \mathbb{E}_{\text{P}_{\pi}^{\kappa}(\cdot\mid s, t)}[\text{V}_{\pi}(\cdot; t+\kappa)] - \mathbb{E}_{\text{P}_{\pi}^{\kappa}(\cdot\mid s', t)}[\text{V}_{\pi}(\cdot; t+\kappa)] \Bigr|}_{\text{Term II}}.
\end{align}

Because $0\le r\le R_{\max}$, Term I is bounded by $R_{\max}\frac{1-\gamma^{\kappa}}{1-\gamma}$. For Term II, note that $\text{V}_{\pi}(\cdot; t+\kappa)$ is bounded by $\frac{R_{\max}}{1-\gamma}$. Using the total variation bound \eqref{eq:short_memory_tv} (which holds uniformly in $t$),
\begin{equation}
\bigl| \mathbb{E}_{\text{P}_{\pi}^{\kappa}(\cdot\mid s, t)}[V_{\pi}] - \mathbb{E}_{\text{P}_{\pi}^{\kappa}(\cdot\mid s', t)}[V_{\pi}] \bigr| \le \|V_{\pi}\|_{\infty} \cdot \bigl\| \text{P}_{\pi}^{\kappa}(\cdot\mid s, t) - \text{P}_{\pi}^{\kappa}(\cdot\mid s', t) \bigr\|_{\mathrm{TV}} \le \frac{R_{\max}}{1-\gamma} \cdot \epsilon.
\end{equation}
Thus we obtain a uniform bound:
\begin{equation}
\bigl| \text{V}_{\pi}(s; t) - \text{V}_{\pi}(s'; t) \bigr| \le R_{\max}\frac{1-\gamma^{\kappa}}{1-\gamma} + \gamma^{\kappa} \frac{R_{\max}}{1-\gamma} \epsilon =: \Delta(\gamma,\kappa,\epsilon).
\label{eq:vdiff_bound}
\end{equation}

\begin{lemma}[Time-Varying Relative Value Homogeneity]
Assume the uniform short-memory property ($\beta_{\pi}(k)\to0$ as $k\to\infty$ uniformly in $t$) and that from every state $s$ at any time $t$, the agent has a probability $p_{\min}>0$ of obtaining a strictly positive expected one-step reward (i.e., there exists $r_{\min}>0$ such that $\mathbb{E}[r_{t} \mid s_t=s] \ge p_{\min} r_{\min}$ for all $s,t$). Define the fleet-average value at time $t$ as $\bar{V}_t = \frac{1}{N}\sum_{i=1}^N \text{V}_{\pi}(s_{i,t}; t)$. Then for any desired relative tolerance $\delta>0$, there exist a discount factor $\gamma\in(0,1)$ and an integer $\kappa$ such that for all agents $i$ and all times $t$,
\begin{equation}
\frac{|\text{V}_{\pi}(s_{i,t}; t)-\bar{V}_t|}{\bar{V}_t} \le \delta.
\label{eq:value_homogeneity_assumption}
\end{equation}
\end{lemma}
\begin{proof}
From \eqref{eq:vdiff_bound}, the maximum deviation between any two state-value functions at the same time $t$ is at most $\Delta(\gamma,\kappa,\epsilon)$. A lower bound on $\bar{V}_t$ can be obtained by noting that from any state, the agent can expect at least $p_{\min} r_{\min}$ immediate reward, and thereafter the discounted sum is non-negative. Hence
\begin{equation}
\bar{V}_t \ge \frac{p_{\min} r_{\min}}{1-\gamma}.
\end{equation}
Consequently,
\begin{equation}
\frac{|\text{V}_{\pi}(s_{i,t}; t)-\bar{V}_t|}{\bar{V}_t} \le \frac{\Delta(\gamma,\kappa,\epsilon)}{p_{\min} r_{\min} / (1-\gamma)} = \frac{R_{\max}}{p_{\min} r_{\min}}\bigl(1-\gamma^{\kappa} + \gamma^{\kappa}\epsilon\bigr).
\end{equation}
Choose $\kappa$ so large that $\epsilon = \beta_{\pi}(\kappa) \le \frac{p_{\min} r_{\min}}{2 R_{\max}}\delta$ and $\frac{1}{\kappa} \le \frac{p_{\min} r_{\min}}{2 R_{\max}}\delta$ (possible because $\beta_{\pi}(\kappa)\to0$ and $\frac{1}{\kappa}\to0$). Set $\gamma = 1 - \frac{1}{\kappa^2}$. For $\kappa\ge2$, we have $\gamma^{\kappa} \ge 1 - \frac{1}{\kappa}$, thus $1-\gamma^{\kappa} \le \frac{1}{\kappa}$. Substituting gives
\begin{equation}
\frac{R_{\max}}{p_{\min} r_{\min}}\left(\frac{1}{\kappa} + \epsilon\right) \le \frac{R_{\max}}{p_{\min} r_{\min}}\left( \frac{p_{\min} r_{\min}}{2 R_{\max}}\delta + \frac{p_{\min} r_{\min}}{2 R_{\max}}\delta \right) = \delta.
\label{eq:example}
\end{equation}
Therefore, for every agent $i$, $|\text{V}_{\pi}(s_{i,t}; t) - \bar{V}_t| \le \delta \bar{V}_t$, completing the proof.
\end{proof}

The lemma establishes that, under realistic mixing conditions (uniform over time) and a mild probabilistic lower bound on obtaining positive rewards, the relative spread among agents' value functions can be made arbitrarily small by choosing the discount factor sufficiently close to $1$. In practice, our system operates in exactly this regime: mixing occurs within minutes, whereas demand evolves over tens of minutes; moreover, any idle vehicle can reach a busy area with positive probability, guaranteeing $p_{\min}>0$. Therefore, for the purpose of algorithm design and analysis, we adopt the following approximation: There exists a time-dependent common baseline $\bar{V}_t$ (shared by all agents but possibly varying slowly with $t$) and a small constant $\delta \ge 0$ such that for every agent $i$ and every time step $t$, Eq. \eqref{eq:value_homogeneity_assumption} holds. Equivalently, $\text{V}_{\pi}(s_{i,t}; t) \approx \bar{V}_t$ up to a small relative error. This property will be used to justify mean‑field approximations and to simplify value estimation in our decentralized coordination scheme. In the remainder of the paper, we simplify the notation by incorporating the time index into the state variable $s_{i,t}$. Thus far, we have established the validity of Eq. \eqref{eq:assumption}.


\subsection{One-Step Advantage Function} \label{sec:ospo}
Here, we provide a detailed derivation of Eq. \ref{eq:marl_advantage}.

\begin{equation}
\begin{aligned}
& \text{A}^{OSPO}_{\pi}(s_{i,t},u_{i,t})  =  \text{Q}_{\pi}(s_{i,t}, u_{i,t}) - \text{V}_{\pi}(s_{i,t}) \\
= & \text{r}(s_{i,t}, u_{i,t}) + \gamma \mathbb{E}_{s_{i,t+1} \sim \text{P}(\cdot|s_{i,t}, u_{i,t})} [\text{V}_{\pi}(s_{i,t+1})] - \text{V}_{\pi}(s_{i,t}) \\
= & \text{r}(s_{i,t}, u_{i,t}) + \gamma \mathbb{E}_{s_{i,t+1} \sim \text{P}(\cdot|s_{i,t}, u_{i,t})} [\bar{V}_{t+1} - \omega(s_{i,t+1})] - \bar{V}_t + \omega(s_{i,t}) \\
= & \text{r}(s_{i,t}, u_{i,t}) + \gamma \bar{V}_{t+1} - \gamma \mathbb{E}_{s_{i,t+1} \sim \text{P}(\cdot|s_{i,t}, u_{i,t})} [\omega(s_{i,t+1})] - \mathbb{E}_{i \in \mathcal{I}, s_{i,t} \sim d^\pi_t(i)} [\text{V}_{\pi}(s_{i,t})]  + \omega(s_{i,t}) \\
= & \text{r}(s_{i,t}, u_{i,t}) + \gamma \bar{V}_{t+1} - \gamma \mathbb{E}_{s_{i,t+1} \sim \text{P}(\cdot|s_{i,t}, u_{i,t})} [\omega(s_{i,t+1})] \\
& - \mathbb{E}_{i \in \mathcal{I}, s_{i,t} \sim d^\pi_t(i), u_{i,t} \sim \pi(s_{i,t}), s_{i,t+1} \sim \text{P}(\cdot|s_{i,t}, u_{i,t})} [\text{r}(s_{i,t}, u_{i,t}) + \gamma \text{V}_{\pi}(s_{i,t+1})]   + \omega(s_{i,t}) \\
= & \text{r}(s_{i,t}, u_{i,t}) + \gamma \bar{V}_{t+1} - \gamma \mathbb{E}_{s_{i,t+1} \sim \text{P}(\cdot|s_{i,t}, u_{i,t})} [\omega(s_{i,t+1})] \\
& - \mathbb{E}_{i \in \mathcal{I}, s_{i,t} \sim d^\pi_t(i), u_{i,t} \sim \pi(s_{i,t})} [\text{r}(s_{i,t}, u_{i,t}))] - \gamma  \mathbb{E}_{i \in \mathcal{I}, s_{i,t+1} \sim d^\pi_{t+1}(i)} [\text{V}_{\pi}(s_{i,t+1})]    + \omega(s_{i,t}) \\
= & \text{r}(s_{i,t}, u_{i,t}) + \gamma \bar{V}_{t+1} - \gamma \mathbb{E}_{s_{i,t+1} \sim \text{P}(\cdot|s_{i,t}, u_{i,t})} [\omega(s_{i,t+1})] \\
& - \mathbb{E}_{i \in \mathcal{I}, s_{i,t} \sim d^\pi_t(i), u_{i,t} \sim \pi(s_{i,t})} [\text{r}(s_{i,t}, u_{i,t}))] - \gamma \bar{V}_{t+1}  + \omega(s_{i,t}) \\
= & \underbrace{\text{r}(s_{i,t}, u_{i,t}) - \mathbb{E}_{i \in \mathcal{I}, s_{i,t} \sim d^\pi_t(i), u_{i,t} \sim \pi(s_{i,t})} [\text{r}(s_{i,t}, u_{i,t}))]}_{\text{Reward Term}} \underbrace{- \gamma \mathbb{E}_{s_{i,t+1} \sim \text{P}(\cdot|s_{i,t}, u_{i,t})} [\omega(s_{i,t+1})] + \omega(s_{i,t})}_{\text{Residual Term} \ \epsilon_{i,t}} \\
\end{aligned}
\end{equation}

Specifically, the residual term satisfies:
\begin{equation}
\begin{aligned}
|\epsilon_{i,t}|
\leq |\omega(s_{i,t})| + \gamma \bigl|\mathbb{E}[\omega(s_{i,t+1})]\bigr|
\leq 2\delta\,\overline{V}_t.
\end{aligned}
\label{eq:residual}
\end{equation}
Consider a scenario in which the fleet-average value $\overline{V}_t$ admits a finite upper bound (a condition automatically satisfied in any finite-horizon setting with episode length $T$). Assume further that the system exhibits sufficiently fast mixing, such that the required mixing lag $\kappa$ is only mildly affected by the horizon length. As shown previously, the relative error $\delta$ can be made arbitrarily small by choosing a discount factor $\gamma$ sufficiently close to $1$ and an appropriate mixing lag $\kappa$. Consequently, one can select these parameters such that the residual term $|\epsilon_{i,t}|$ becomes arbitrarily small and can therefore be safely neglected in the advantage approximation.


\section{Algorithm} \label{sec:alg}

The detailed algorithm of our ST-GRPO/OSPO for ride-sharing system is illustrated in Algorithm \ref{alg:grpo}.

\begin{algorithm}[htbp]
\caption{ST-GRPO/OSPO for Ride-Sharing System}
\label{alg:grpo}
\begin{algorithmic}[1]
\REQUIRE Policy network $\pi_{\theta}$ with shared parameters $\theta$
\REQUIRE Best policy checkpoint $\pi_{\phi}$ (initialized as $\pi_{\theta}$)
\REQUIRE Environment simulator, Reward function (Eq. \eqref{eq:reward_func})
\FOR{episode $= 1$ to $M$}
    \STATE Initialize experience buffer $\mathcal{D} \gets \emptyset$
    \STATE Reset environment, get initial states $\{s_{i,0}\}_{i=1}^n$, $s^o_0$
    \FOR{time step $t = 0$ to $T$}
        \FOR{each agent $i \in \mathcal{I}$}
            \STATE Compute action probabilities (Eq. \eqref{eq:action})
        \ENDFOR
        \STATE Solve assignment problem via BiPartite matching (Eq. \eqref{eq:ILP}), getting action $U_t$
        \STATE Execute assignment $U_t$, observe rewards $\{r_{i,t+1}\}_{i=1}^n$
        \STATE Store transition $(\{s_{i,t}\}, s^o_t, U_t, \{r_{i,t+1}\}, \{s_{i,t+1}\}, s^o_{t+1})$ in $\mathcal{D}$
    \ENDFOR
    \FOR{epoch $= 1$ to $H$}
        \STATE Sample batch $\mathcal{B} \sim \mathcal{D}$
        \STATE Compute group advantage $\mathcal{A}_t$
        \STATE Update policy parameters $\theta$ by maximizing Eq. \eqref{eq:GRPO}
    \ENDFOR
    \IF{current policy $\pi_{\theta}$ outperforms $\pi_{\phi}$}
        \STATE Update best checkpoint: $\phi \gets \theta$
    \ENDIF
\ENDFOR
\end{algorithmic}
\end{algorithm}

\section{Experiment Details} \label{sec:detail}

\subsection{Evaluation Metrics} \label{sec:metric}
The evaluation metrics used in this paper include:
\begin{itemize}[left=0pt]
    \item \textbf{Served Rate:} The proportion of confirmed trips relative to the total number of trip requests initiated by customers.
    \vspace{0.1cm}
    \item \textbf{Delivery Time:} The total time taken to serve a trip from origin to destination.
    \vspace{0.1cm}
    \item \textbf{Detour Time:} The additional time spent on delivery beyond the minimum delivery time, which is defined as the time required if the vehicle serves only this trip without bundling others.
    \vspace{0.1cm}
    \item \textbf{Pickup Time:} The waiting time for customers between trip confirmation and the arrival of the vehicle at the trip origin.
    \vspace{0.1cm}
    \item \textbf{Confirmation Time:} The waiting time for customers from the moment they initiate a trip request to when the platform assigns the trip to a vehicle.
\end{itemize}

Specifically, for delivery time and detour time, only completed orders are counted, as these metrics are uncertain for unfinished orders.

\subsection{Introduction of Comparative Methods}
To illustrate the efficiency of our methods, we compare them with several representative ride-sharing approaches from various categories:
\begin{itemize}[left=0pt]
    \item \textbf{DeepPool \cite{al2019deeppool}:} DeepPool is one of the earliest MARL-based ride-sharing works, based on IDDQN \cite{IQL,van2016deep}, in which each agent trains its Q-network separately. To align it with our experimental scenario, we replace the CNN used in original paper with an MLP similar to our methods (without the softmax).
    \vspace{0.1cm}
    \item \textbf{BMG-Q \cite{hu2025bmg}:} To achieve better cooperation among agents, BMG-Q is also based on IDDQN but utilizes Graph Attention Network (GAT) \cite{velickovic2017graph} to capture the relationships among neighboring agents.
    \vspace{0.1cm}
    \item \textbf{HIVES \cite{hao2022hierarchical}:} Based on QMIX \cite{rashid2020monotonic}, HIVES tries addressing the curse of dimensionality problem by proposing a novel hierarchical mixing structure.
    \vspace{0.1cm}
    \item \textbf{Enders et al. \cite{enders2023hybrid}:} Based on Multi-Agent Soft Actor Critic (MASAC) \cite{haarnoja2018soft}, Enders et al. propose allowing agents to choose whether to accept assigned orders, which helps the system serve more valuable orders instead of treating all orders equally.
    \vspace{0.1cm}
    \item \textbf{CEVD \cite{bose2023sustainable}:} CEVD modifies the Value Decomposition (VD) paradigm \cite{sunehag2018value} from CTDE to CTCE. Specifically, it combines the Q-values of each agent with those of their neighbors to create a new type of Q-value, similar to the motivation behind BMG-Q.
    \item \textbf{Hoppe et al. \cite{hoppe2024global}:} Building on the work of \cite{enders2023hybrid}, a method based on COMA \cite{foerster2018counterfactual} is developed that utilizes a mixture of global and local rewards. Specifically, the global reward encourages cooperation, while the local reward enhances training efficiency.
\end{itemize}
For fairness, we modify the state space and reward function to align closely with those used in this paper.

\subsection{Detailed Experiment Results} 

\begin{figure}[htbp]
\centering 
\includegraphics[width=0.7\textwidth]{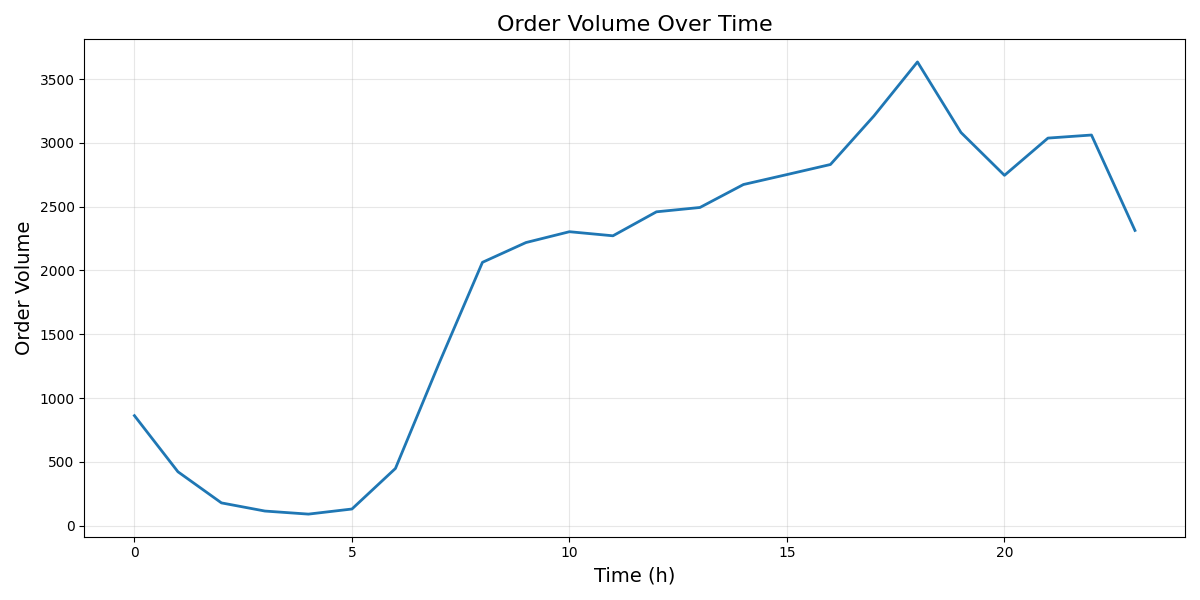}
\caption{Order Temporal Distribution}
\label{fig:order}
\end{figure}

\begin{figure*}[t!]
\centering 
\includegraphics[width=\textwidth]{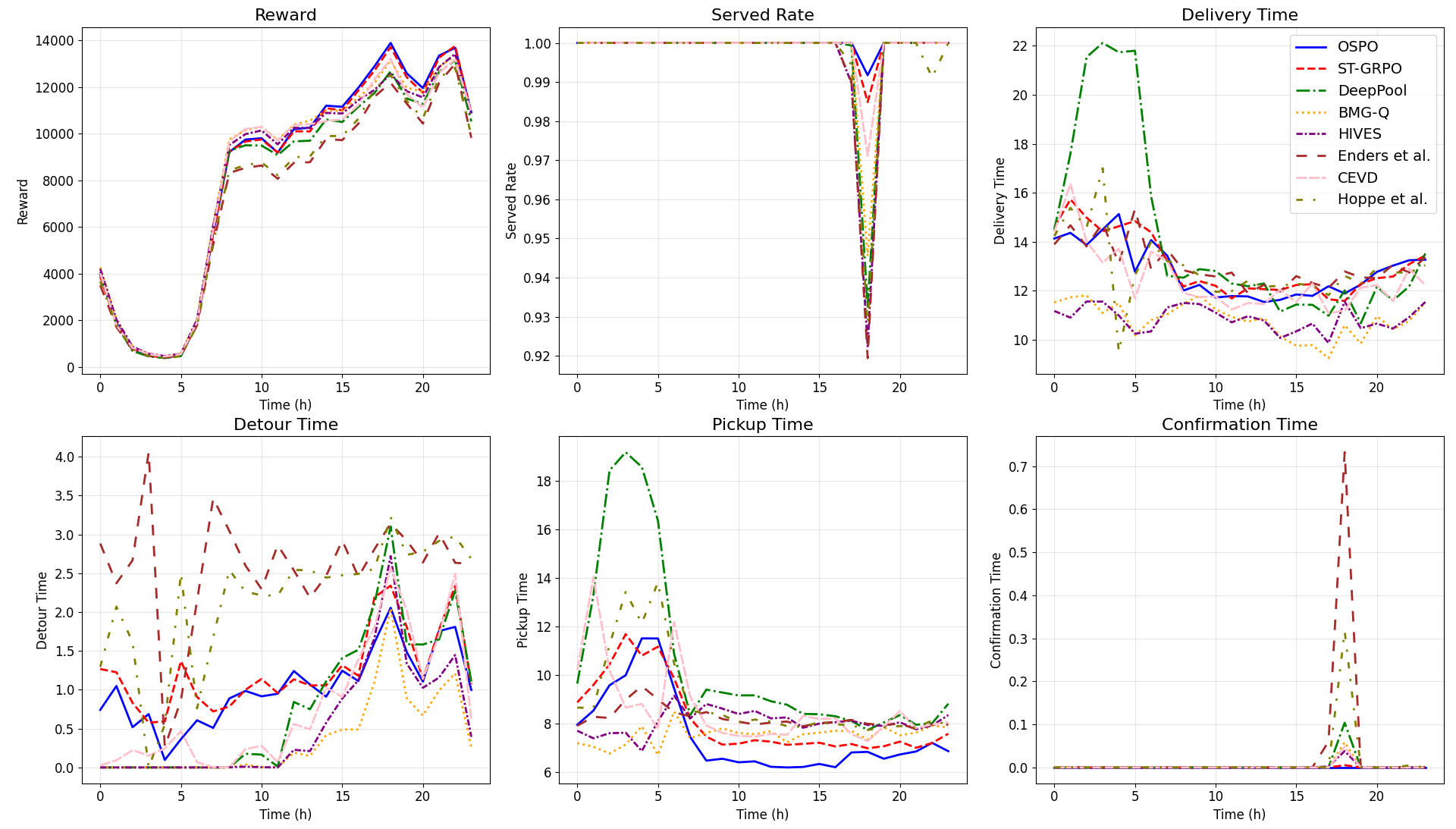}
\caption{Experiment Results in Testing Set}
\label{fig:test}
\end{figure*}

In Fig. \ref{fig:order} and \ref{fig:test}, we illustrated the order amount and the metrics of different methods under each hour in the whole day. First, it's intuitive that the trend of reward is similar to the order volume, since more orders can bring more profits to platform. Then we notice before 8 a.m., when the order volume is very low, the overall performance of different methods is similar, since the idle vehicles are sufficient. However, after that, the BMG-Q and HIVES first illustrate slightly advantage during 8 a.m. to 1 p.m. and then are outperformed by our ST-GRPO and OSPO methods as the order volume further increases after 1 p.m.. Overall, our ST-GSPO and OSPO mainly outperform others by optimizing the pickup time and confirmation time, leading to the highest order service rate in the on-peak hour 6 p.m..

\begin{figure*}[t!]
\centering 
\includegraphics[width=\textwidth]{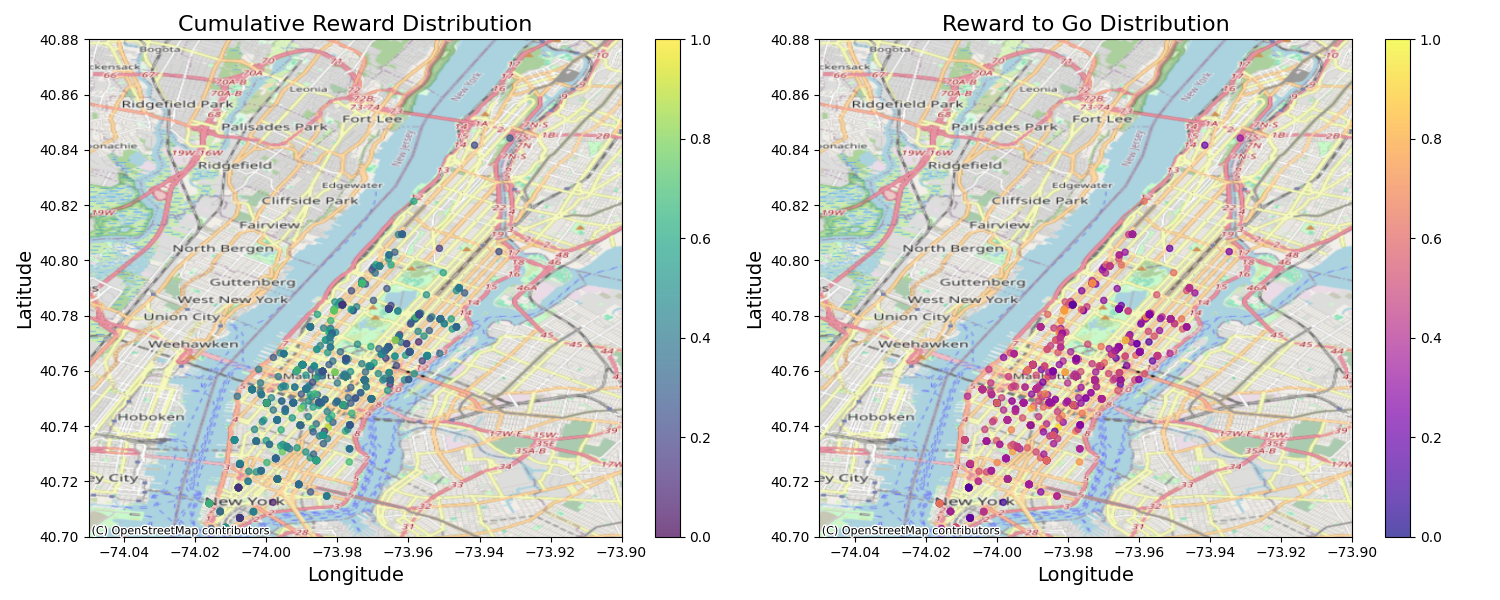}
\caption{The relationship between initial location and metrics about cumulative reward and reward-to-go.}
\label{fig:assumption}
\end{figure*}

Additionally, to further prove our homogeneity assumption, we illustrate the relationship among initial state (location) and metrics regarding cumulative reward (without discount factor $\gamma$) and reward-to-go (with discount factor $\gamma$) of our method. The results suggest that in our method, the state does not significantly influence the long-term normalized cumulative reward and reward-to-go. As shown in Fig. \ref{fig:assumption}, this result supports our homogeneity assumption from a localization perspective. Specifically, the Coefficient of Variation (CV) of the reward‑to‑go (V-value estimation) is only 13.45\%, indicating relatively low divergence. 

\subsection{Ablation Study} \label{sec:ablation}

In the ablation study, we primarily investigate the effects of:
\begin{itemize}[left=0pt]
    \item \textbf{Reward Normalization:} We evaluate the effect of reward normalization in both ST-GRPO and OSPO by testing model performance after removing this module.
    \vspace{0.1cm}
    \item \textbf{Deviation Punishment Regulation:} We examine the impact of including deviation punishment ($\zeta_{t}-\zeta_{t-1}$ in Eq. \eqref{eq:new_advantage3}) to determine whether it helps to maintain homogeneous property (Eq. \eqref{eq:assumption}).
    \vspace{0.1cm}
    \item \textbf{KL Regulation Term:} We assess the effect of using a KL regulation term ($\text{KL}(\pi_\theta \| \pi_{ref})$ in Eq. \eqref{eq:GRPO}) to illustrate whether the historical best policy can prevent model divergence.
\end{itemize}

\begin{figure*}[t!]
\centering 
\includegraphics[width=\textwidth]{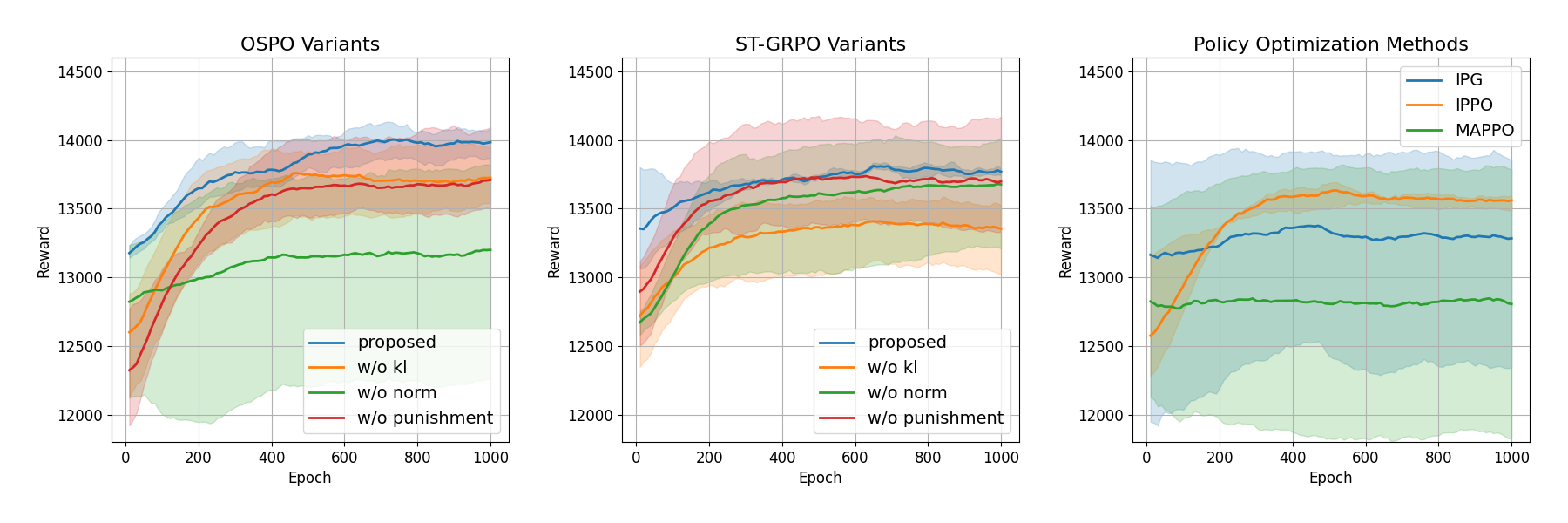}
\caption{Training Process of Ablation Study}
\label{fig:ablation}
\end{figure*}


Additionally, we compare our methods with conventional policy optimization methods, including IPPO, MAPPO, and IPG. As shown in Fig. \ref{fig:ablation}, we notice that the proposed ST-GRPO and OSPO outperform all variant versions, illustrating the efficiency of the proposed modules. For the policy optimization methods, we observed that only IPPO demonstrates relatively good performance (even when compared to our proposed methods), while IPG struggles with high derivation and low sample efficiency. Additionally, MAPPO suffers from the curse of dimensionality in the centralized critic network, resulting in failure to converge.

Additionally, we observe that OSPO demonstrates overall better performance than ST-GRPO. This advantage arises from the nature of our ride-sharing scenario, where the number of served orders may vary slightly among agents and episodes. For instance, an agent may receive a newly assigned order at the last step of the simulation, resulting in a relatively higher reward for that episode. While this variation is negligible over the long term, it becomes more pronounced within the confines of our 30-minute simulation when calculating the advantage based on the average reward in Eq. \eqref{eq:grpo_advantage3}. In contrast, OSPO exhibits greater robustness, as it relies solely on one-step rewards and does not encounter this issue.

\subsection{Expanded Experiment} \label{sec:expand}

\begin{table*}[t!]
    \caption{Performance of Different Methods in Queens, New York City \cite{TLCData}}
    \centering
    \begin{adjustbox}{width=1.00\textwidth}
    \begin{tabular}{l@{\quad}|ccccccc}
        \toprule
        \textbf{Method} & \textbf{Reward} & \textbf{Service Rate} & \textbf{Delivery Time} & \textbf{Detour Time} & \textbf{Pickup Time} & \textbf{Confirmation Time} \\
        \midrule
        \textbf{DeepPool} & 5222.85 & 0.64 & \textbf{11.24} & 1.84 & 12.30 & \textbf{0.21} \\
        \textbf{BMG-Q}  & 5362.00 & 0.66 & 9.63 & 1.08 & 12.98 & 0.27  \\
        \textbf{HIVES}  & 3560.80 & 0.60 & 8.30 & \textbf{0.36} & 14.67  & 0.41 \\
        \textbf{Enders et al.} & 4543.68 & 0.61 & 10.41 & 0.85 & 13.39 &  2.25  \\
        \textbf{CEVD}  & 4388.83 & 0.62 & 11.61 & 1.33  & 13.74  & 0.29 \\
        \textbf{Hoppe et al.} & 4728.93 & 0.63 & 9.78 & 0.53 & 13.45 & 2.18 \\
        \midrule
         \textbf{ST-GRPO} & 5365.89 & 0.64 & 10.56 & 1.86 & 12.21 & \textbf{0.12} \\
         \textbf{OSPO} & \textbf{5521.74} & \textbf{0.67} & 11.37 & 2.21 & \textbf{11.88} & 0.21 \\
        \bottomrule
    \end{tabular}
    \end{adjustbox}
    \label{tab:new}
\end{table*}

To further demonstrate the generalization of our method, we conduct experiments using High Volume For-Hire Vehicle (FHV) trip data from Queens, New York City \cite{TLCData}. Unlike the Manhattan data used in our previous experiment, Queens is a larger area with a lower order density. This difference in order distribution can complicate the process of determining which orders to bundle together for vehicle sharing. In this scenario, we select the data from 19:00 to 19:30 on July 17, 2024, which includes 2,024 valid orders and 500 vehicles. The detailed experimental results are presented in Table \ref{tab:new}. 

The results demonstrate that our proposed ST-GRPO and OSPO methods continue to exhibit superior performance, primarily by optimizing pickup and confirmation times. Specifically, OSPO significantly outperforms the other methods, including ST-GRPO. This advantage arises because, in a more challenging and dynamic scenario (due to the enlarged area), the ST-GRPO method is more severely impacted by errors associated with the bounded time horizon, whereas OSPO leverages a one-step reward to mitigate this shortcoming.

\section{Discussions} \label{sec:discussion}

\subsection{Application Conditions of the Proposed Methods} \label{sec:condition}

In this section, we summarize the conditions under which our methods are applicable and explain their practical implications in real-world AV ride-sharing systems. First, both ST-GRPO and OSPO require the following two conditions:

\textbf{Condition I: Homogeneity (Identical AV Fleet):} All agents must share the same internal attributes, state space, action space, and reward function, and the task must be fully cooperative. In the context of ride-sharing systems, this means that all AVs adopt the same vehicle model, the system policy is non-discriminatory, and each agent's objective is to maximize the global cumulative reward.


\textbf{Condition II: Short Memory (Rapid Mixing):} The system must be insensitive to initial conditions after a moderate time horizon. Concretely, this requires that the operational region be sufficiently connected and that the order distribution be diverse enough to relocate vehicles across different areas within a short period. It means that orders have a roughly equal probability of driving each AV to arbitrary locations.

Moreover, our methods rely on a problem setup condition: the discount factor $\gamma$ should be sufficiently large. This condition is used to establish the V-value homogeneity property, which depends on the objective construction rather than the system dynamics themselves. Furthermore, OSPO introduces an additional application condition:

\textbf{Condition III: Bounded V-Value (e.g. Finite Horizon):} In OSPO, the condition for replacing the reward-to-go with a single-step reward is that the V-value is upper bounded. This condition is satisfied when the horizon is finite and the system mixes quickly, which is practical in ride-sharing systems where the running time is finite and newly assigned orders can rapidly relocate AVs to new positions.

\subsection{Why OSPO is not Myopic}

At first glance, OSPO may appear to be a myopic method because it relies only on single-step rewards. In this section, we explain why it is not myopic, from theoretical, intuitive, and experimental perspectives.

\textbf{(i) Theoretical perspective:} The advantage function of OSPO (Eq. \eqref{eq:new_advantage2}) is derived from the standard advantage function with the objective of maximizing long-term cumulative reward (Eq. \eqref{eq:GRPO}), which is fundamentally different from the myopic objective of maximizing single-step rewards. The reason OSPO can simplify the original advantage function into a single-step form is due to the three conditions discussed in the previous subsection, which arise from the inherent properties of the AV ride-sharing system.

\textbf{(ii) Intuitive perspective:} What OSPO essentially does is to find a good policy $\pi$ shared by all agents (AVs) to increase the probability of beneficial matches while decreasing that of detrimental ones (Eq. \eqref{eq:new_advantage2}). Since all agents share nearly identical V-values at each step, the relative quality of an action can be assessed locally from its immediate contribution. This can be interpreted as optimizing a matching policy shared by all agents, which differs from myopic methods that simply maximize the global single-step reward greedily. Moreover, compared to conventional independent MARL solutions, both OSPO and ST-GRPO consider all agents together in the advantage function, thereby enabling implicit cooperation.

\textbf{(iii) Experimental perspective:} We observe that when the normalization module is removed from OSPO, model performance degrades significantly, as shown in Fig. \ref{fig:ablation} and Appendix \ref{sec:ablation}. This occurs because removing the normalization term reduces the advantage in Eq. \eqref{eq:new_advantage2} to a pure reward term $\text{r}(s_{i,t}, u_{i,t})$, making OSPO degenerate into a purely myopic method. This leads to an optimization objective that is entirely different from that of our OSPO, which is derived with theoretical guarantees. Furthermore, as reported in \cite{sun2022optimizing, sun2024optimizing}, myopic solutions generally fail to achieve V-value homogeneity and exhibit pronounced spatial disparities, whereas our OSPO does achieve it, as evidenced in Fig. \ref{fig:assumption}.

\subsection{Comparison between GRPO, ST-GRPO, and OSPO}
To illustrate the differences between our proposed methods (ST-GRPO and OSPO) and the original GRPO \cite{GRPO}, we summarize them in Table \ref{tab:difference}.

\begin{table}[t]
\centering
\caption{Comparison of GRPO \cite{GRPO}, ST-GRPO, and OSPO}
\label{tab:difference}
\begin{tabularx}{\columnwidth}{@{} l X X X @{}}
\toprule
& \textbf{GRPO \cite{GRPO}} & \textbf{ST-GRPO} & \textbf{OSPO} \\
\midrule
\textbf{Application Scenario} 
& LLM post-training 
& Cooperative MARL with Condition ~I--II 
& Cooperative MARL with Condition I--III \\
\midrule
\textbf{Theoretical Guarantee} 
& $\times$ 
& \checkmark 
& \checkmark \\
\midrule
\textbf{Training Samples} 
& Multiple trajectories from the same initial state 
& Single trajectory of multiple agents 
& Single-step samples from multiple agents \\
\midrule
\textbf{Normalization Term} 
& Reward-to-go over multiple trajectories 
& Reward-to-go over multiple agents 
& Single-step reward over multiple agents \\
\bottomrule
\end{tabularx}
\end{table}

\subsection{Limitations and Future Work} \label{sec:limitation}

Our method is specifically tailored to homogeneous AV ride-sharing systems with fast mixing speed. While this design choice trades off generality for training efficiency and strong empirical performance, the underlying conditions are readily met in many practical deployments. For instance, many ride-sharing systems operate with identical AV fleets within well-connected metropolitan areas (e.g., differently colored taxis in Hong Kong and New York), which naturally leads to fast mixing. Experimental results also demonstrate the robustness of our method across different scenarios under similar experimental settings as previous work. As a next step, it is worthwhile to explore the applicability of our method to tasks that share similar conditions with our AV ride-sharing setting. It is also worth investigating how to extend our method to more general ride-sharing scenarios, such as those with heterogeneous fleets.

\end{document}